# Identifying the exterior image of buildings on a 3D map and extracting elevation information using deep learning and digital image processing

**Donghwa Shon [1], Byeongjoon Noh [2] and Nahyang Byun [1*]**

[1] Department of Architecture, Chungbuk National University, Cheongju, Chungbuk 28644, S.Korea; dhshon@cbnu.ac.kr(D.S.); nhbyun@chungbuk.ac.kr(N.B.)
[2] Applied Science Research Institute, Korea Advanced Institute of Science and Technology, Daejeon 34141, S.Korea; powernoh@kaist.ac.kr
* Correspondence: nhbyun@chungbuk.ac.kr

**Abstract:** Despite the fact that architectural administration information in Korea has been providing high-quality information for a long period of time, the level of utility of the information is not high because it focuses on administrative information. While this is the case, a three-dimensional (3D) map with higher resolution has emerged along with the technological development. However, it cannot function better than visual transmission, as it includes only image information focusing on the exterior of the building. If information related to the exterior of the building can be extracted or identified from a 3D map, it is expected that the utility of the information will be more valuable as the national architectural administration information can then potentially be extended to include such information regarding the building exteriors to the level of BIM(Building Information Modeling). This study aims to present and assess a basic method of extracting information related to the appearance of the exterior of a building for the purpose of 3D mapping using deep learning and digital image processing. After extracting and preprocessing images from the map, information was identified using the Fast R-CNN(Regions with Convolutional Neuron Networks) model. The information was identified using the Faster R-CNN model after extracting and preprocessing images from the map. As a result, it showed approximately 93% and 91% accuracy in terms of detecting the elevation and window parts of the building, respectively, as well as excellent performance in an experiment aimed at extracting the elevation information of the building. Nonetheless, it is expected that improved results will be obtained by supplementing the probability of mixing the false detection rate or noise data caused by the misunderstanding of experimenters in relation to the unclear boundaries of windows.

## 1. Introduction

The Korean government implemented the “Act on the Promotion of the Provision and Use of Public Data” in 2013, which discloses data owned by the government and the local governments to the public for the purpose of economic development and improvements in quality of life [1]. Information owned by government institutions is provided through the Public Data Web Portal, which, as of August 2020, includes 22 government-focused datasets and approximately 38,396 individual pieces of data. An increasing number of successful business projects use public information disclosed by the private sector [2].

In regard to urban architecture, 2,752 cases of administrative information are provided, including buildings, houses, land, utility, urban planning, transportation, real estate, infrastructure, and weather, not to mention that the government continues to make effort to expand and upgrade the scope of the data collected.

However, among the data regarding urban architecture, there is little real demand for information related to architecture as it is too high in proportion to administrative information, compared to high-quality data construction. This is because architectural administration information primarily comprises licensing information, building ledgers, closure and cancellation ledgers, and energy use and information such as building entities, actions, location, date, size (area, number of floors, etc.), type (use, structure, etc.), and energy use is limited for the private sector.

In the field of construction, there is expected to be a need for discussing public information in the scope of disclosure, as there are various challenges such as copyright and security issues of buildings. Architectural administration information needs to be gradually developed into the data on a BIM(Building Information Modeling) level to improve the value of information utility.

BIM-level public data are applied to all architectural processes, including administration, structure, construction, maintenance, and energy, on the grounds of form-based information. These data need to be developed along with smart architecture and urban technology using digital twin methods that build real-time information in both directions.

Because of the government's efforts, buildings in the future are likely to be built in the form of BIM data. However, measures to develop architectural administration information regarding existing buildings at BIM levels are still needed.

As the first step, a method is required to build building exterior information in linkages that are not recorded in existing architectural administration information. This corresponds to most of the buildings on the national level, and a step-by-step advancement plan that can collect and build information in a uniform fashion by minimizing time, cost, and labor is needed.

The use of deep learning technology can be considered to extract building elevation as a possible way to build information regarding building exteriors using the images of the building exteriors on three-dimensional(3D) maps. On major Korean web portals, image information regarding architecture and urban exteriors has been built as high-quality data in recent years, including street views and 3D maps. Particularly, both the resolution quality of 3D maps and deep learning technology for identifying and categorizing image information is expeditiously improving, suggesting the technical possibility to subdivide and build architectural and urban exterior components by applying the two factors to existing high-quality image information.

For this reason, the current study aims to propose a deep learning technology model that identifies and extracts building exterior information using a 3D map and to provide methodology that assesses the validity of this technology.

## 2. Advances in image identification and extraction technology

### *2.1 3D maps*

Domestic and foreign private IT corporations such as Google and Naver provide map services including street views and aerial satellite photos at the horizontal level as well as 3D map services that show the exteriors of architecture and cities in three dimensions. 3D maps provide users with a more directly immersive urban experience. In addition, the resolution and sophistication of maps are improving as autonomous drones or automobiles are being developed. 3D maps differ from maps simply using satellite images as 3D mapping is a method of photographing the exteriors of architecture and city from several angles using drones or aircraft and synthesizing the photographed images in three dimensions based on coordinate values to create a map. The government has provided two-dimensional (2D) and 3D spatial information (physical and logical spaces as well as properties belonging to figures) connected to public administration information by means of a spatial information open platform (Vworld), and a 3D shape model is being created based on the architectural arrangement form and floor heights. Moreover, 3D spatial information is being built by applying images taken from several angles in relation to the shape model via aerial photographic survey [3]. In particular, the Seoul Metropolitan Government has constructed a 3D map (S-Map) of the entire 605km$^2$ of Seoul and features better accuracy in location information and in resolution than the existing 3D spatial information platform [4]. An advantage of S-Map is that the image is easily extracted with no obstacles covering the exterior of the buildings.

As such, with the rapid development of 3D maps, users are provided with convenience and experiences, but these are only used as visual image data for buildings. It can be argued that information regarding the exteriors of buildings being built as a dataset will be advantageous for the purpose of reproduction or utilization of the information

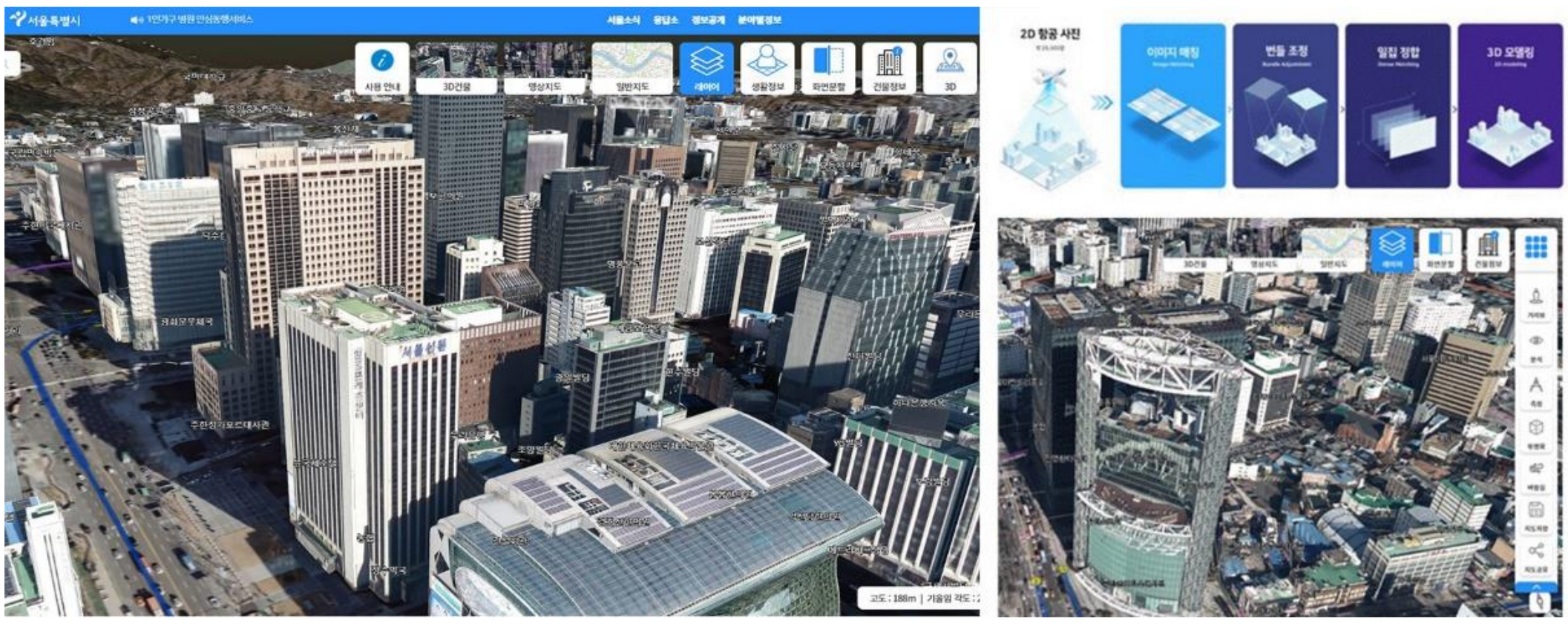

**Figure 1.** Seoul Metropolitan Government S-Map screen

*2.2 Advances in image processing and deep learning technology*

A primary technology for image processing is computer vision (CV) technology. In the 2000s, CV was rapidly improved and was primarily used for the identification, division, tracking, mapping, estimation, and exploration of objects, with the development of large-scale data and artificial intelligence technology due to provision via the Internet. Based on knowledge-base methodology, it primarily focused on feature extraction methodology such as contour detection, corners, and key-points within the image. The application of the technology is divided into image processing in terms of removing noise and emphasizing features and a background analysis aimed at extracting or identifying information. Imaging processing technology based on CV was able to detect objects. However, it was not put into practice until the early 2000s as its utility was limited in terms of the classification of the objects.

After 2010, image processing technology and research have reached a stage of practical utility in many fields, such as autonomous cars, fingerprints, facial recognition, and robot control, with the rapid development of artificial intelligence technology. Arguably, that it has not only extracted the features of the images but has also led to a technique of classifying the extracted features after deep learning technology was applied is a major shift. The importance of data preprocessing and conversion processes such as spatial domain conversion, geometric conversion, and frequency conversion has been reduced, while the detection and classification of objects and motion detection have been made possible through image inputs. Based on CNN(Convolutional Neuron Networks), deep learning technology for image processing is being developed into platforms such as YOLO and Detectron based on several object detection models such as Fast and Faster-R. These platforms provide pre-trained models and image data of a high capacity for commonly used objects such as people, vehicles, chairs, and buildings.

Deep learning-based image identification technology is currently being used in many fields. In medical fields, deep learning-based image processing techniques are used to analyze chest X-ray images, MRI images and ophthalmic images and to develop software that quickly and accurately diagnoses various diseases and aids in effective treatment. Google has developed a system that can diagnose diabetic retinopathy using CNN technique-based Optogram photographic analysis, and Lunit INSIGHT has developed a medical support software system for lung disease and breast cancer diagnosis [5-6]. In the fields of transportation and security, accidents are prevented by tracking and categorizing the movement of objects using closed-circuit television (CCTV) images. As a result, systems were developed to forecast traffic volumes of cars, taxis, bicycles, and other vehicles on roads based on deep learning technology using image data and to create accident predictions by analyzing the movements of pedestrians and vehicles [7-9].

Prior to the application of deep learning-based technology, any attempts to identify the exteriors of buildings were aimed at recognizing building openings. Several approaches were developed, such as deciding whether the opening is glass through RGB value colors in elevated areas within the images and detecting entrances using lines, colors, and textures [10-12]. The image identification technology to which deep learning technology is applied targets images of building exteriors; in particular, elevation was studied for the purpose of identifying the primary components constituting the building or detecting structural combinations. What differentiates this from previous studies prior to deep learning technology development is that identification elements are categorized along with algorithms that extract features from the images. These studies primarily focus on extracting building outlines, elevation characteristics, and outer wall materials [13-15]. It can be said that these studies are similar to previous studies wherein they identify the components of the elevation; however, they are limited in that this includes shapes, elements, and materials. The current study presents a method of identifying elements in buildings as multifold as possible by appropriately distinguishing image identification and deep learning techniques.

### *2.3 Deep learning technology and a video detection deep learning platform*

Shortcomings including slow learning time, data supply and demand problems, and overfitting of existing artificial neural networks prior to the 2000s were solved, which led to the hidden layer of existing artificail neural networks being developed into a deep neural network (DNN) comprising multiple layers. Just as conventional neural networks, DNNs comprise input layers, hidden layers, and output layers and is able to model complex non-linear relationships. DNNs have been recently developed into different forms including recurrent neural networks (RNNs) in natural language processing and convolutional neural networks (CNNs) in computer vision. CNNs are primarily used for image identification as artificial neural networks to imitate visual processing methods, whereas RNNs were designed for the purpose of processing time series data such as audio, sensors, and characters.

Thus, major deep learning platforms for object detection in images are CNN-based. These platforms are being developed into platforms such as YOLO and Detectron based on several object detection models, such as Fast and Faster-R. These platforms provide pre-trained models such as people, vehicles, chairs, buildings, and related high-capacity image data.With Tensorflow, a Python-based deep learning open-source package provided by Google, rsearch can be conducted on several data mining, machine learning, and deep learning models as well as enable people to build large visualizations of models designed via data flow graphs based on GPU(Graphics Processing Unit) and CPU(Central Processing Unit). In addition to Tensorflow, platforms such as PyTorch and Caffe2 can be used. You Only Look Once (YOLO) is a deep learning model that speciailizes in detecting objects in images. YOLO improved issues regarding the computational speed of existing R CNN-series models, which allowed one network to extract features, create bouding boxes, and categorize classes simultaneously [16].

## 3. A model for identifying exterior images and extracting elevation information using deep learning and image processing techniques

The current study selected a neighborhood with a high concentration of high-rise buildings in Jongno-gu, Seoul, as the case subject site to identify the exterior images of buildings and extract information; any buildings that met the criteria were categorized as high-rise buildings (See Figure 2.). This is not only because the images consisting of high-rise buildings are high in resolution but also because it is necessary to establish a model that identifies components within an architectural type frame of high-rise buildings to allow a framework to be established for application to other building types as well. Here, the model for detecting building elevations is proposed by applying deep learning technology. The model largely comprises 1) an image extraction unit, 2) an image preprocessing unit, 3) a building elevation recognition unit, and 4) a data management unit. The model structure is illustrated in Figure 3.

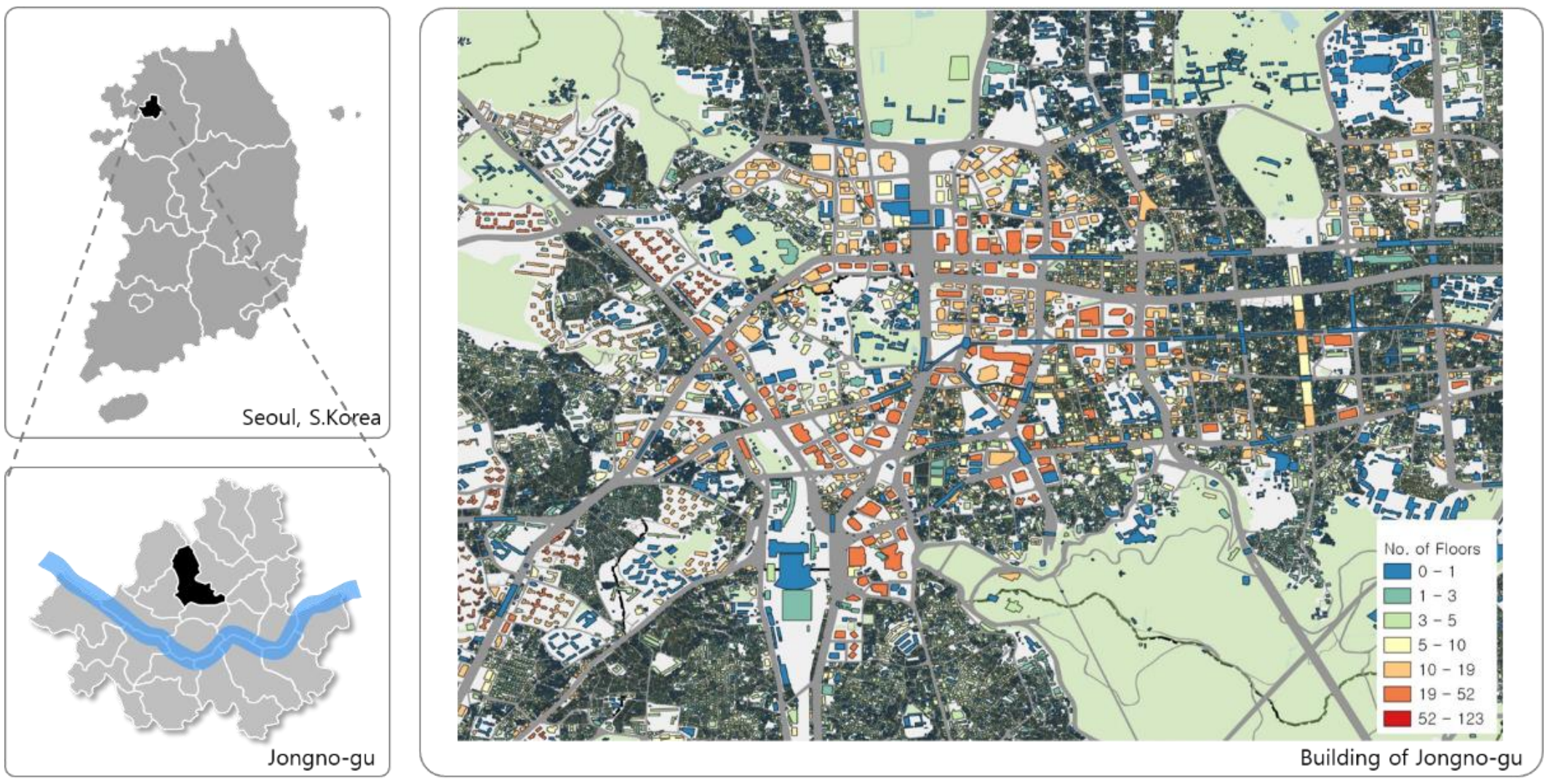

**Figure 2**. Analysis target area around Jongno-gu

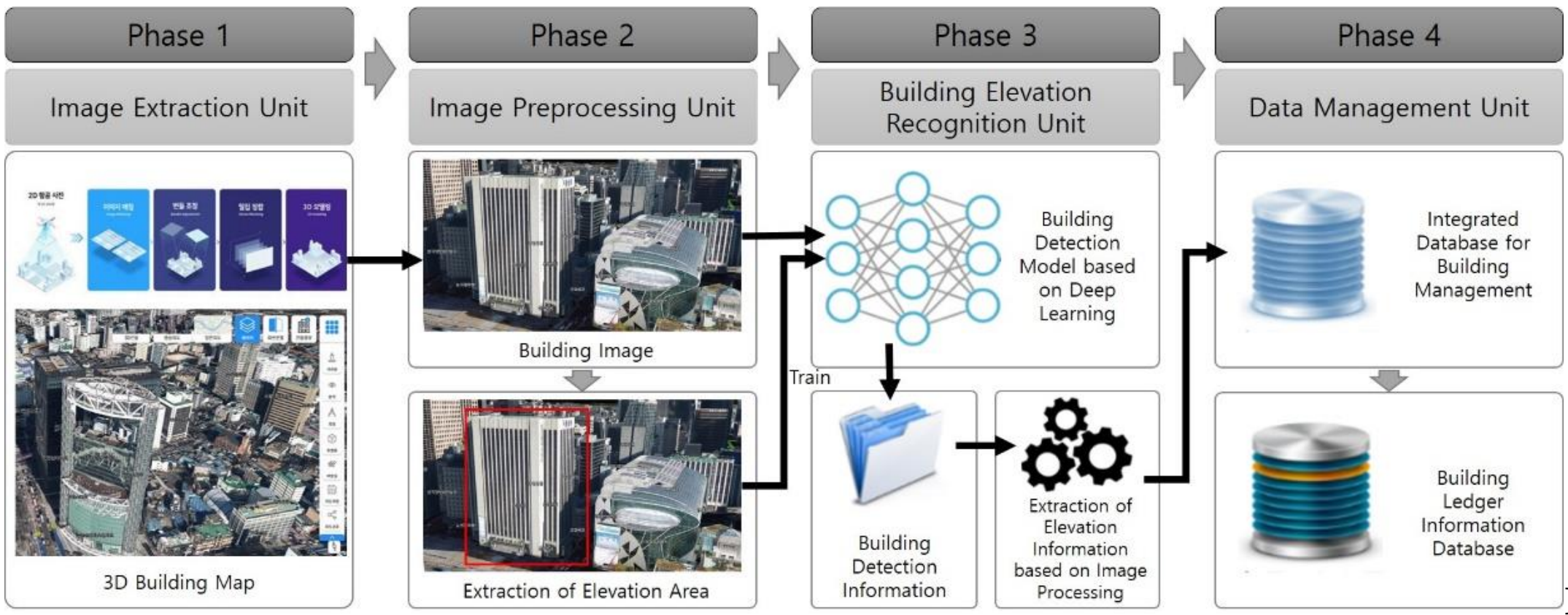

**Figure 3.** A model for identifying exterior images and extracting elevation information using deep learning and image processing techniques

First, a 3D building map image is used to identify the images of the building exteriors and extract elevation information. The image preprocessing unit sets the region of interest (RoI) on the elevation of the building to train the deep learning model from the acquired image. The building elevation recognition unit detects the building exteriors using a deep learning model, wherein the images of building exteriors are learned, and extracts building elevation information from the building that was identified. The identified information related to building elevations is managed in an integrated database for building management, which can also be used to manage buildings by combination with the building ledger information database.

### *3.1 Method used to acquire building 3D map image data*

This section explains a method of acquiring images of building exteriors from 3D building maps. First, the current study used 3D building maps based on the information provided by S-Map [4]. S-Map provides user-friendly 3D-type building maps and enables users to set several layers, including administrative divisions, a state basic district system, road name information, subway information, and building names as well as different map resolutions such as high, normal, and low resolution. Additionally, it provides the viewpoint of the map in the form desired by the user via altitude and tilt angle settings (see Figure 1).

S-Map supports its own application programming interface (API) in general; however, it does not support APIs that extract 3D building images such as those that the current study intends to use. For this reason, the current study implemented a macro program that automatially moves and stores a screen via AutoIt software to acquire an image of a 3D builidng [17]. It is easy to perform repetitive tasks by automating the mouse and keyboard movements on AutoIt software. This software was used to extract approximately 500 elevation images of buildings in Jongno-gu, Seoul. Considering that distortion occurs as the result of the nature of the image in extracting the elevation of buildings, we extracted elevation as we focused the point of view as frontally as possible. The extraction altitude and inclination angle were set to 114m and 14°, respectively.

### *3.2 Method used to preprocess images to train the deep learning model*

This section explains an image processing method to train the deep learning model using the 3D building image acquired by the image extraction unit. Generally, when a deep learning model is being trained to detect objects, coordinates and images for a section corresponding to an object area within an image called a bounding box are used as well. In general, the bounding box has the shape of a square and is shown in the form of left upper coordinates (minx, miny) and right lower coordinates (maxx, maxy). Figure 4 illustrates an example of displaying an object area in an image as a bounding box.

The current study set a bounding box for the building areas from the extracted image information to train the deep learning building detection model. The settings for the bounding box were set manually for all images for the model learning dataset as well as the test dataset, and the VGG Image Annotator (VIA) open-source software [18] was utiized to extract bounding box coordinates. When training the deep learning model, metadata or annotation containing object information in the image along with images required for learning are necessary. Metadata contain information including image paths, sizes, and bounding box coordinates and are generally stored in the format of either JavaScript Object Notation (JSON) or Extensible Markup Language (XML). VIA open-source software supports object ranging and metadata extraction for training the model. Figure 4 is the result of setting bounding boxes in 3D building images using the VIA open-source software. In this study, a bounding box was set up to recognize the elevation and windows of the building from the 3D building images.

The current study extracted image metadata in the format of JSON, the structure of which is as shown in Table 1.

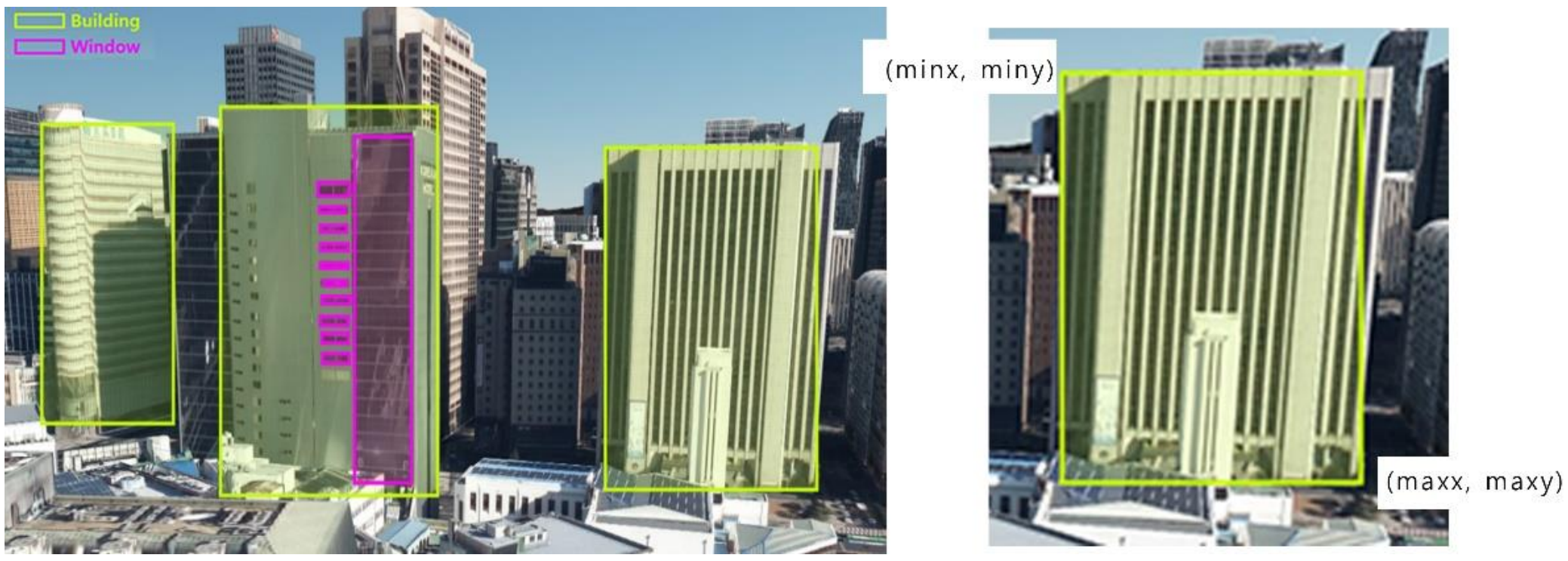

**Figure 4.** An example of bounding box and settings in the 3D building image

**Table 1.** Object labeling metadata structure for learning

| **Key values** | **Description** | **Item values (example)** | **Data format** |
|---|---|---|---|
| metadata | Image metadata information | {...} | DICTIONARY |
| └ filename | Image file name | 'test_image.png' | STRING |
| └ size | File size | 3202282 | INT |
| └ regions | Bounding box properties information | [...] | LIST |
| └ shape_attributes | Object size information | {...} | DICTIONARY |
| └ name | Object area information | 'rect' | STRING |
| └ x | Bounding box top left-hand corner x coordinate | 643 | INT |
| └ y | Bounding box top left-hand corner y coordinate | 123 | INT |
| └ width | Bounding box width | 359 | INT |
| └ height | Bounding box height | 371 | INT |
| └ region_attributes | Object attribute information | {...} | DICTIONARY |
| └ class | Object class | 'building' | STRING |
| file_attributes | Image file property information. | {...} | DICTIONARY |
| .... (continued) | | | |

Table 2 is an example of a JSON file used for learning. In the given example, building elevation information was extracted from the 3D image files of 'image1.png' and 'image2.png', with the file sizes being approximately 3MB (megabytes). The 'region' key value includes attribute information such as a bounding box size, location, class information, and similar component information. The 'image1.png' file contains information regarding two buildings and one window as well as location information (top left hand corner x and y coordinates, length, and height) for each bounding box. Similarly, it can be interpreted that 'image2.png' file contains only two pieces of building information.

**Table 2.** An example of JSON including labeled object information

```
{ 'image1_info' : { 'filename' : 'image1.png' ,
                    'size' : 3202282,
                    'regions' : [{ 'shape_attributes' : { 'name' : 'rect' ,
                                                          'x' : 643,
                                                          'y' : 123,
                                                          'width' : 359,
                                                          'height' : 371 },
                                   'region_attributes' : { 'class' : 'building' }},
                                 { 'shape_attributes' : { 'name' : 'rect' ,
                                                          'x' : 175,
                                                          'y' : 415,
                                                          'width' : 30,
                                                          'height' : 23 },
                                   'region_attributes' : { 'class' : 'window}},
                                 { 'shape_attributes' : { 'name' : 'rect' ,
                                                          'x' : 689,
                                                          'y' : 1048,
                                                          'width' : 290,
                                                          'height' : 454 },
```

'region_attributes' : { 'class' : 'building' }}],
'file_attributes' : {}},
'image2_info' : { 'filename' : 'image2.png' ,
'size' : 390471,
'regions' : [{ 'shape_attributes' : { 'name' : 'rect' ,
'x' : 125,
'y' : 213,
'width' : 259,
'height' : 471 },
'region_attributes' : { 'class' : 'building' }},
{ 'shape_attributes' : { 'name' : 'rect' ,
'x' : 689,
'y' : 1048,
'width' : 290,
'height' : 454 }
'region_attributes' : { 'class' : 'building' }}],
'file_attributes' : {}}
}

*3.3 Building elevation recognition unit*

This section explains a method for training a deep learning model and one for recognizing the elevation of buildings using preprocessed image information and image metadata. This section will also partially explain how various building information can be acquired using image processing techniques from the recognized building elevation.

3.3.1 Deep learning model-based method to recognize building elevation

Prior to the development of deep learning models, existing object detection technology was based on in-image algorithms such as the scale-invariant feature transform (SIFT) and speeded-up robust features (SURF) or on machine extraction algorithms including K-nearest neighbor (KNN) and Support Vector Machine (SVM) to extract the main features in the image using mathematical approximations. Afterward, with the improvements in deep learning technology, the speed and accuracy of detecting objects in images have rapidly improved. The major algorithms for object detection in images are as follows: DetectorNet [19], SPP Net [20], VGG Net [21], Regions with Convolutional Neural Network (R-CNN) [7], Fast/Faster R-CNN [23-24], You Only Look Once (YOLO) [25-28], and Mask R-CNN [29].

This study aims to learn 3D building images using Fast R-CNN, which is an improved model of R-CNN, and then use them for detection and extraction of building elevation information. In general, the R-CNN model extracts a 'Regional Proposal' in the image using the selective search algorithm and then extracts features in the image using the CNN model in each candidate area. Classification is carried out to categorize objects using SVM based on the extracted features. In the Fast R-CNN that emerges, a separate Regional Proposal Network (RPN) is applied to extract feature maps. Additionally, the location of the object can be extracted by performing a bounding box regression operation, which has higher mAP (mean average precision). The structure of Faster R-CNN is illustrated in Figure 5.

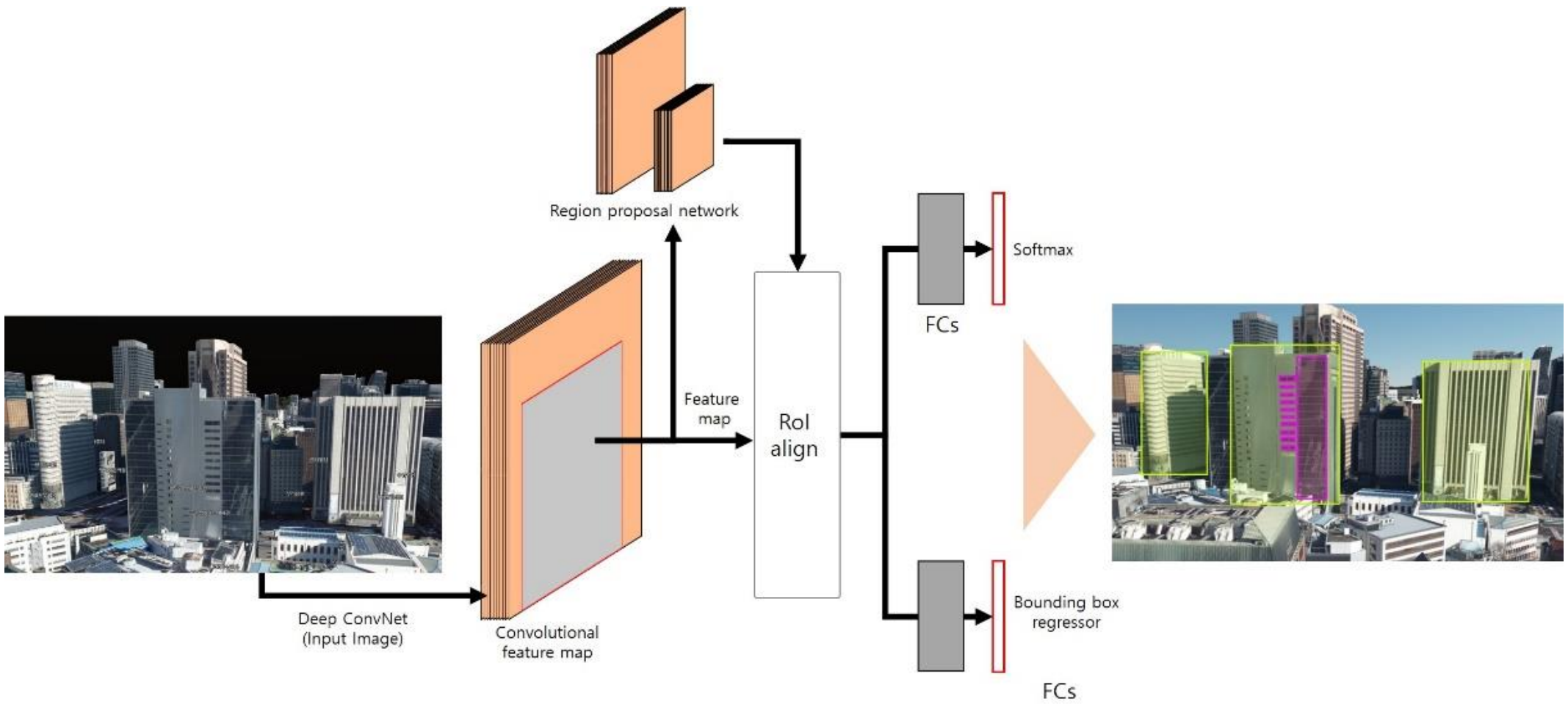

**Figure 5**. Faster R-CNN model structure

When learning, label data including images for learning, building elevation bounding box information, and window bounding box information in images are used together in the JSON file format for data configuration for 3D building image learning.

In this study, the Detectron2 platform [30-31] provided by Facebook AI Research is used to learn given learning data (image and label).

As the result of the learning as follows, we can acquire a learning model wherein the deep learning model is finally up-to-date. Figure 6 illustrates the result of the 3D building image tested in the learned model. The results show that high-rise buildings are well-detected, but windows are not due to the distortion of the buildings and the image size. Section 4 outlines the result of the model learning and its performance.

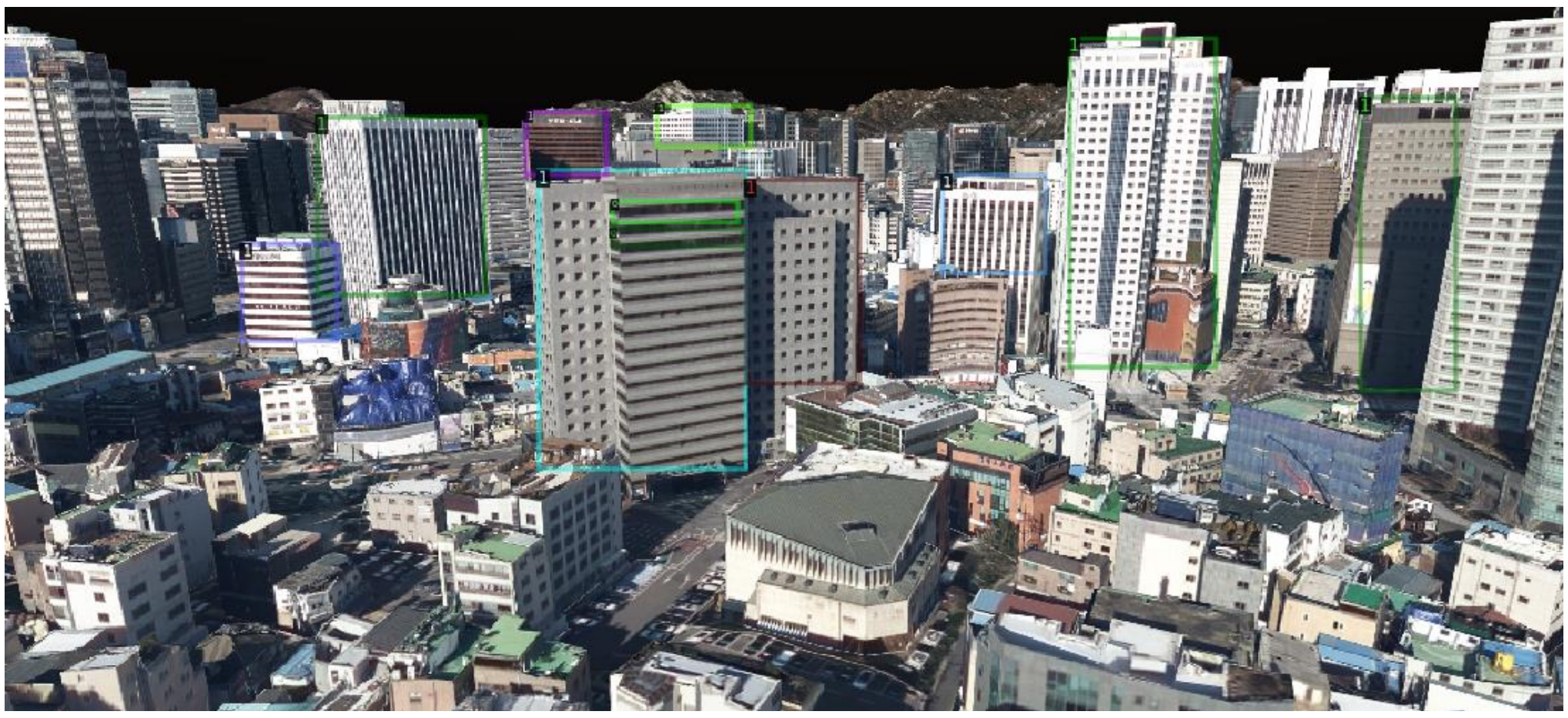

**Figure 6**. The test result image of the model that learned the 3D building image

3.3.2 Method of extracting building elevation information based on image processing technology

This section outlines the method of extracting elevation information of a building using the image processing technique based on the results presented through the deep learning model. The exteriors of the building and its elevation vary, and the 3D building exterior image used in the current study in particular is not clear in terms of image quality and shape. For this reason, the study only conducted an experiment to distinguish windows from walls among the exterior and elevation information of the building.

The method of classifying windows and walls is divided into a simple classification method and a detailed classification method.

Simple classification method for windows: In the simple classification method for windows, only the existence or absence of windows is identified based on the elevation of the building. As previously mentioned, windows can also be recognized through learning window areas when learning an elevation image of buildings. Generally, the window areas exist within the elevation area of the building. Nevertheless, due to the noises in the image area, it is sometimes detected outside the elevation area (see Figure 7). To prevent this, the study compared the locations between the elevation areas and the window areas of the building, which improved detection accuracy. As a result, a set of window areas ($w_j$) included in the building elevation area ($B_i$) can be acquired. The equation for comparison between the detected building elevation area ($B_i$) and the detected window area ($w_j$) is as follows:

$$W_i = \{w_x \mid threshold \leq |w_x \cap B_i| \leq |w_x|; x = 1,2,\ldots,m\}; i = 1,2,\ldots,n$$

where $|A|$ is a size of area $A, m$ is the number of detected windows,

and $n$ is the number of detected buildings

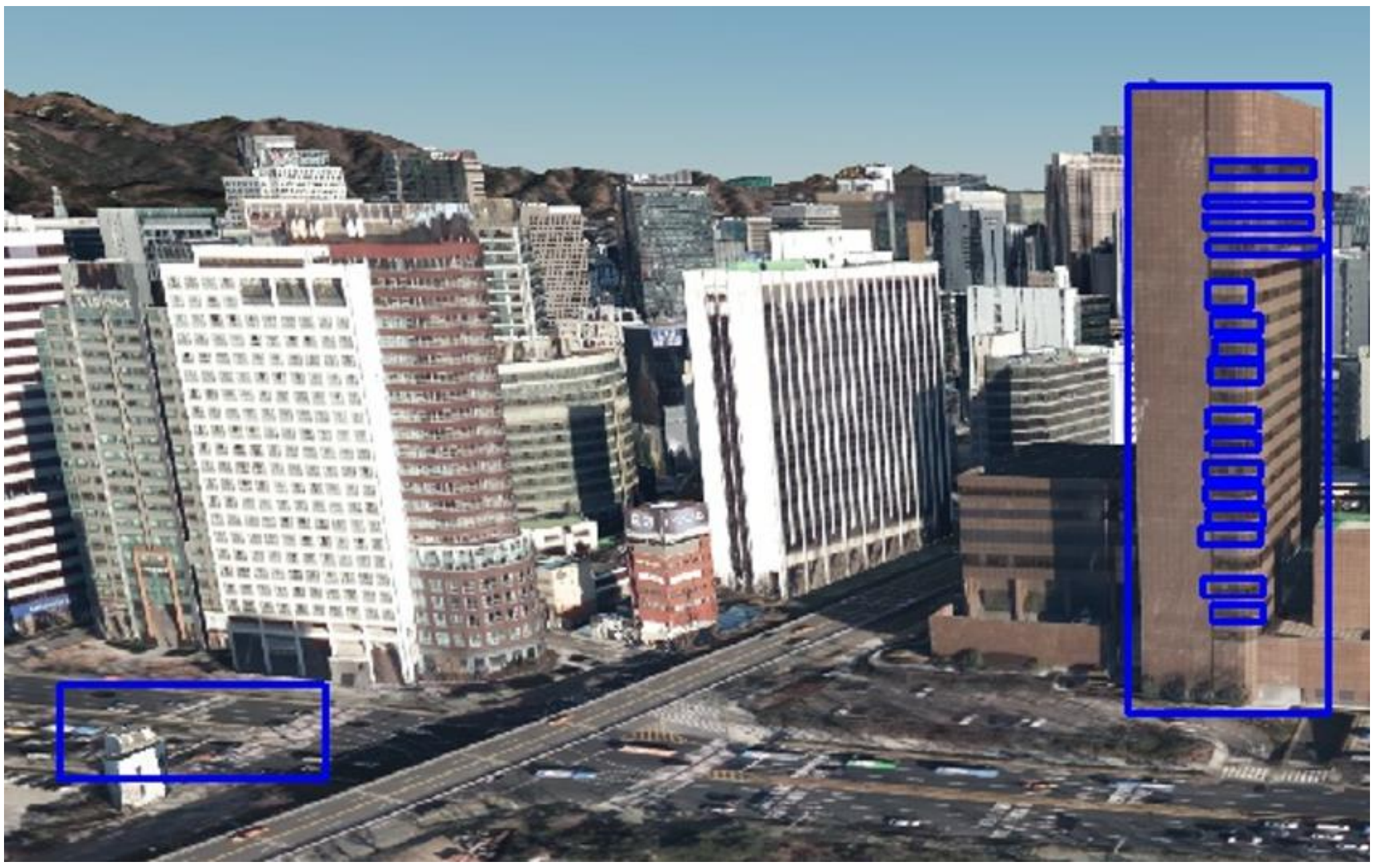

**Figure 7.** An example of cases in which the window area in the elevation of buildings is properly detected (right) and where it is not (left)

When there is a window area within the elevation area of the building, the building is categorized as having windows; otherwise, it is categorized as a building with no windows. This is used as part of the preprocessing in the detailed classification method as described below.

Detailed classification method for windows: The detailed method is mainly twofold, and information regarding the following two factors is extracted: the type of elevation window and 2) the ratio of elevation windows. The classified contents for each detail are as follows:

- Type of elevation window: front curtain wall, repeated single windows, others (mixed windows, etc.)
- Elevation window ratio: ~25%, ~50%, ~75%, ~100%

Among the types of elevation windows, the front curtain wall is when the window covers the entire building, and the single window repetition is when the window is repeated by floor and pattern. Finally, the “other” category includes mixed windows and half-height horizontal windows.

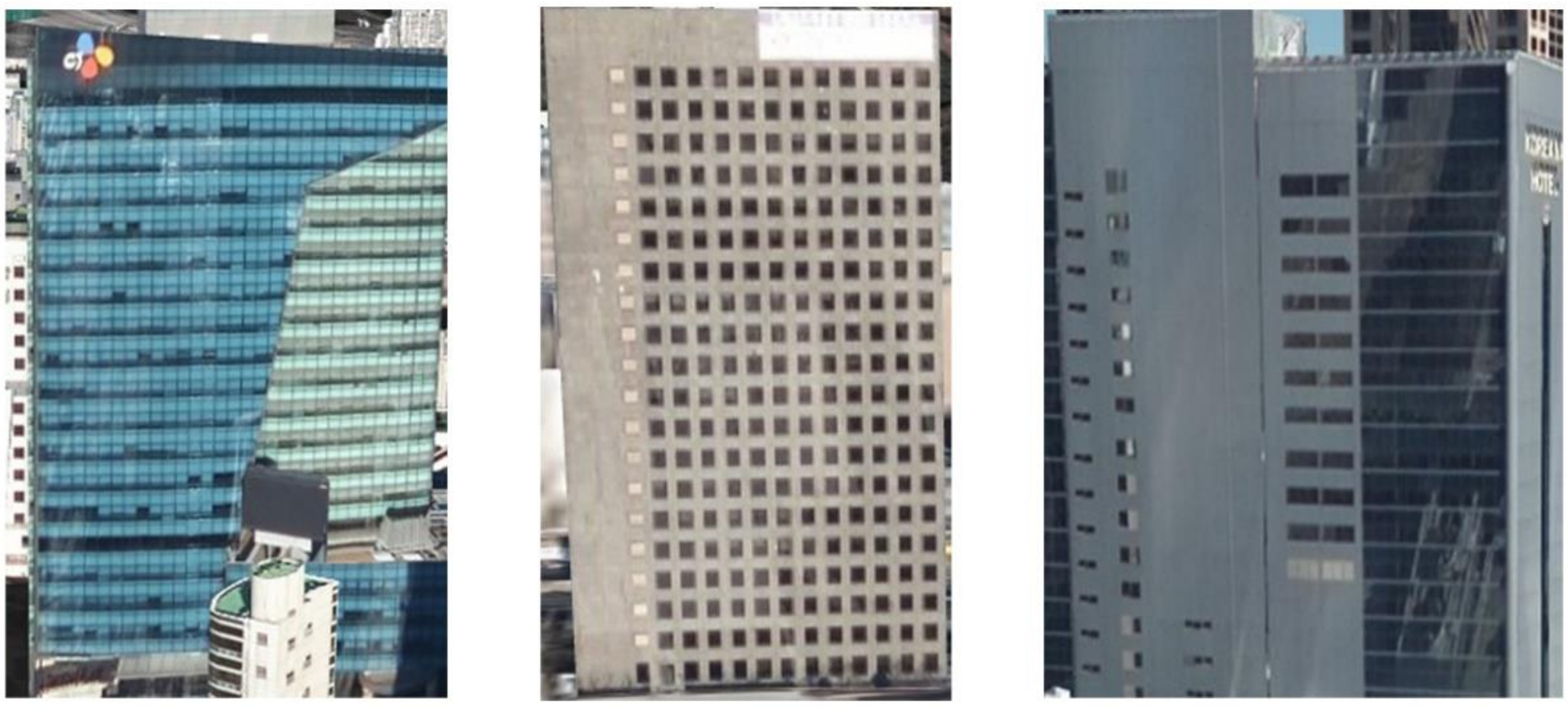

(a) (b) (c)

**Figure 8.** 3D buildings detected on the building map: (a) Front curtain wall building, (b) Repeated single window building, (c) "Other" (mixed windows)

The distinction between the front curtain wall and the repeated single windows can be acquired by using the number of window areas recognized in the elevation area of the building. Buildings in the "Other" category refer to the form to which both factors can be applied (mixed windows). When extracting the exteriors of buildings based on a deep learning model, the distinction of buildings and windows was carried out in advance. For this reason, they are to be classified through the spacing of window areas within an elevation area of a building. As previously mentioned, provided that the size of the area between the building elevation area ($B_i$) and the window area ($w_x \in W_i$) within the building elevation area is similar or equal, the building is classified as a front curtain wall building. If two or more window areas are detected, the building is classified as a repeated single window building. Buildings that are not classified as the aforementioned two building types are classified as "Other." The classification function $D$ to detect the type of elevation windows is as follows:

$$D = \begin{cases} \text{front curtain wall,} & |B_i| - |w_x| \leq threshold;\, i = 1, 2, \ldots, n;\; x = 1, 2, \ldots, |W_i|,\; w_x \in W_i \\ \text{repeated single windows,} & x \geq 2 \\ \text{others,} & otherwise \end{cases}$$

Next, as for the classification of the elevation window ratio, the ratio of the size of the window area to the size of the elevation area of buildings is utilized. The window ratio is classified into four categories, ~25%, ~50%, ~75%, ~100%, and the window ratio $R$ can be acquired using the following equation:

$$R = 100 * \sum_{x=1}^{|W_i|} |w_x| \Big/ |B_i| \;\; ; \;\; w_x \in W_i;\, i = 1,2, \ldots, n$$

A simple classification method for walls: The current study used RGB (Red, Green, Blue) channel values to distinguish colors in the image so that the color of the wall is distinguished. First, RGB channel values in the elevation area of the building were extracted, excluding the window areas, and colors were classified as black, maroon, silver, orange, green, and blue. The following RGB channel values for each color are defined as reference channel values:

- Black: $K(r, g, b) = (0, 0, 0)$
- Maroon: $R(r, g, b) = (90, 0, 0)$
- Silver: $S(r, g, b) = (192, 192, 192)$
- Orange: $Y(r, g, b) = (255, 127, 0)$
- Green: $G(r, g, b) = (0, 255, 0)$

- Blue: $B(r,g,b) = (0,0,255)$

The average of the RGB channel values for the parts excluding the window area among the elevation areas of buildings was used and compared to the RGB channel values for each color as presented above. The color of the elevation of buildings is determined by comparing the average RGB channel value and the reference RGB channel value in the corresponding area. The function for comparing RGB channel values is as follows [32]:

$$\underset{i;\, U_i \in U}{\operatorname{argmin}}\, l(U,A)\,;\, U = \{K,R,S,Y,G,B\}$$

$$l(U,A) = \sqrt{\left((512 + rm) * r_i * r_i \gg 8\right) + 4 * g_i * g_i + (767 - rm) * b * b) \gg 8};rm = (r_i + \bar{r})/2$$

The average RGB channel value of the corresponding building elevation is $A(\bar{r},\bar{g},\bar{b}) = \left(\frac{\sum_{i=1}^{n} r_i}{n}, \frac{\sum_{i=1}^{n} g_i}{n}, \frac{\sum_{i=1}^{n} b_i}{n}\right)$, where $n$ refers to the number of pixels in the detection region. Figure 9 illustrates the RGB channel and the assigned color of each building elevation.

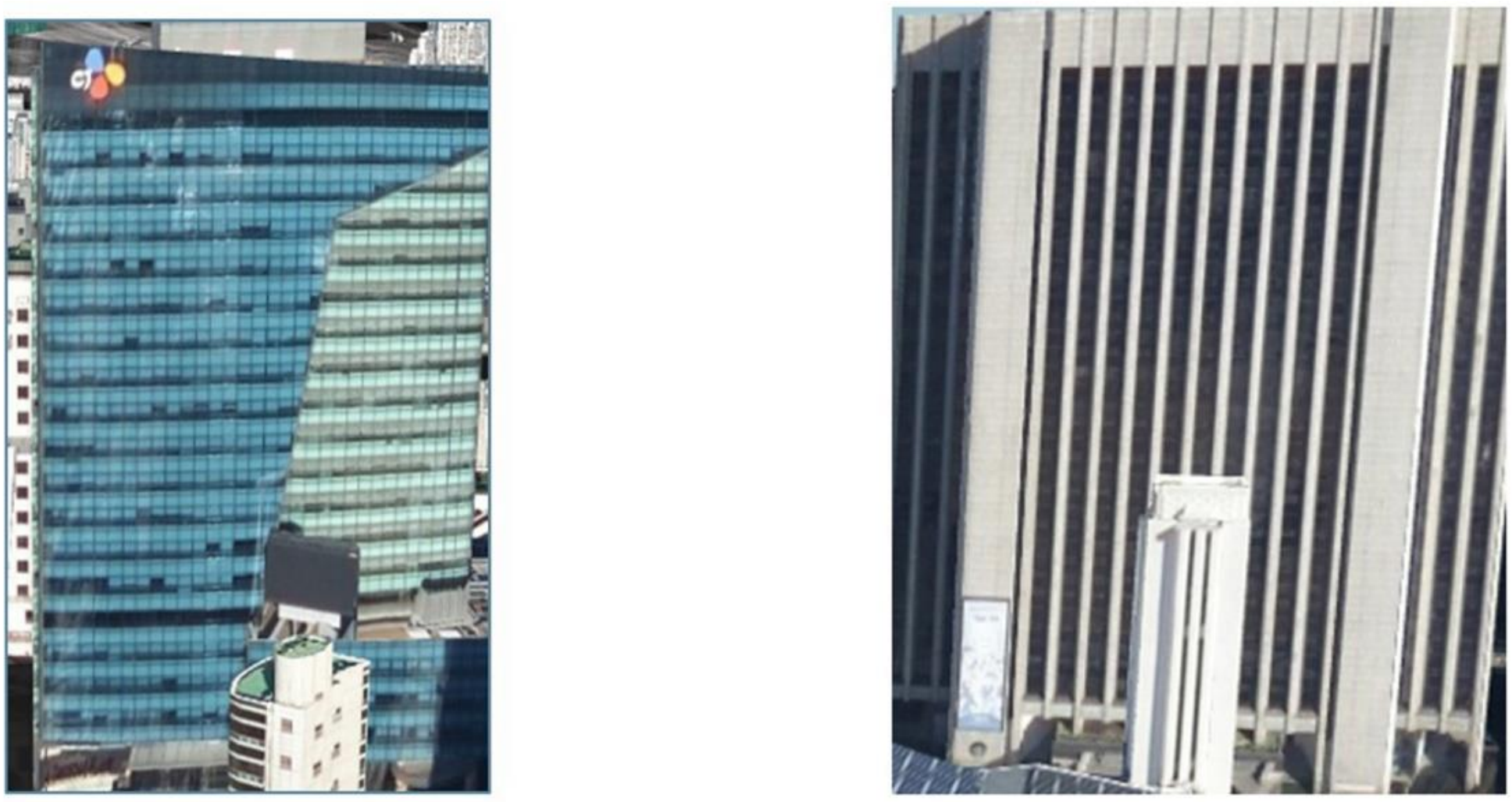

**Figure 92.** Example of allocating elevation colors based on RGB building channels

The detailed method to classify the walls: The detailed method aims to classify the materials (glass, cement, etc.) of the elevation of the building. To recognize and classify the material of the elevation of the building, materials can be labeled again and classified by observing the detected elevation image of the building. This can be conducted in a manner similar to processes in which the elevation and windows of the building are detected. The current study did not conduct the detailed classification method of the walls separately. Nonetheless, in light of the results acquired after detecting the elevation and windows, there are sufficient reasons to consider that classification can potentially be carried out separately. In addition, materials that are not clearly distinguished with the naked eye, such as cement, bricks, and tiles, could be considered together using the elevation color classification method for buildings that is used in the simple classification method.

## 4. Experiment and results

### *4.1 Experiment design*

This section explains the experimental performance plan for identifying the elevation image of buildings to which deep learning technology was applied as described in Section 3 and verifying the performance of the elevation information extraction methodology. First, the experimental process and results for verifying the performance of the model to recognize building elevation using deep learning technology is described. Next, a process for extracting building elevation information and an experimental result thereof are outlined using an image prcoessing technique from the extracted image.

*4.2 The results of building elevation recognition using the deep learning model.*

This section explains the results of building elevation recognition using the deep learning model. The current study used the Fast R-CNN model in the Detectron2 environment to recognize the elevation of the building from the elevation image of the building. The total number of acquired images is approximately 800, and the number of windows is approximately 900. Among the acquired images, approximately 388 (70%) images were used for learning, and approximately 166 (30%) images were used for the test. The building elevation image used here utilized the image of high-rise buildings in Seoul. Regarding the hyperparameters for learning, the learning rate was set to 0.0005, and the maximum number of iterations was set to 5,000 times. Additionally, by using the early stopping function of learning, the learning is stopped when there is no further reduction in loss by iteration. In the current study, iteration occurred approximately 1200 times and was stopped after learning. Figure 10 illustrates the graph showing losses by iteration. Generally, the loss function utilizes the Root Mean Squared Error (RMSE), and it is meaningful to be aware that the model is learning from the decreasing trend of loss values with iteration as opposed to absolute values.

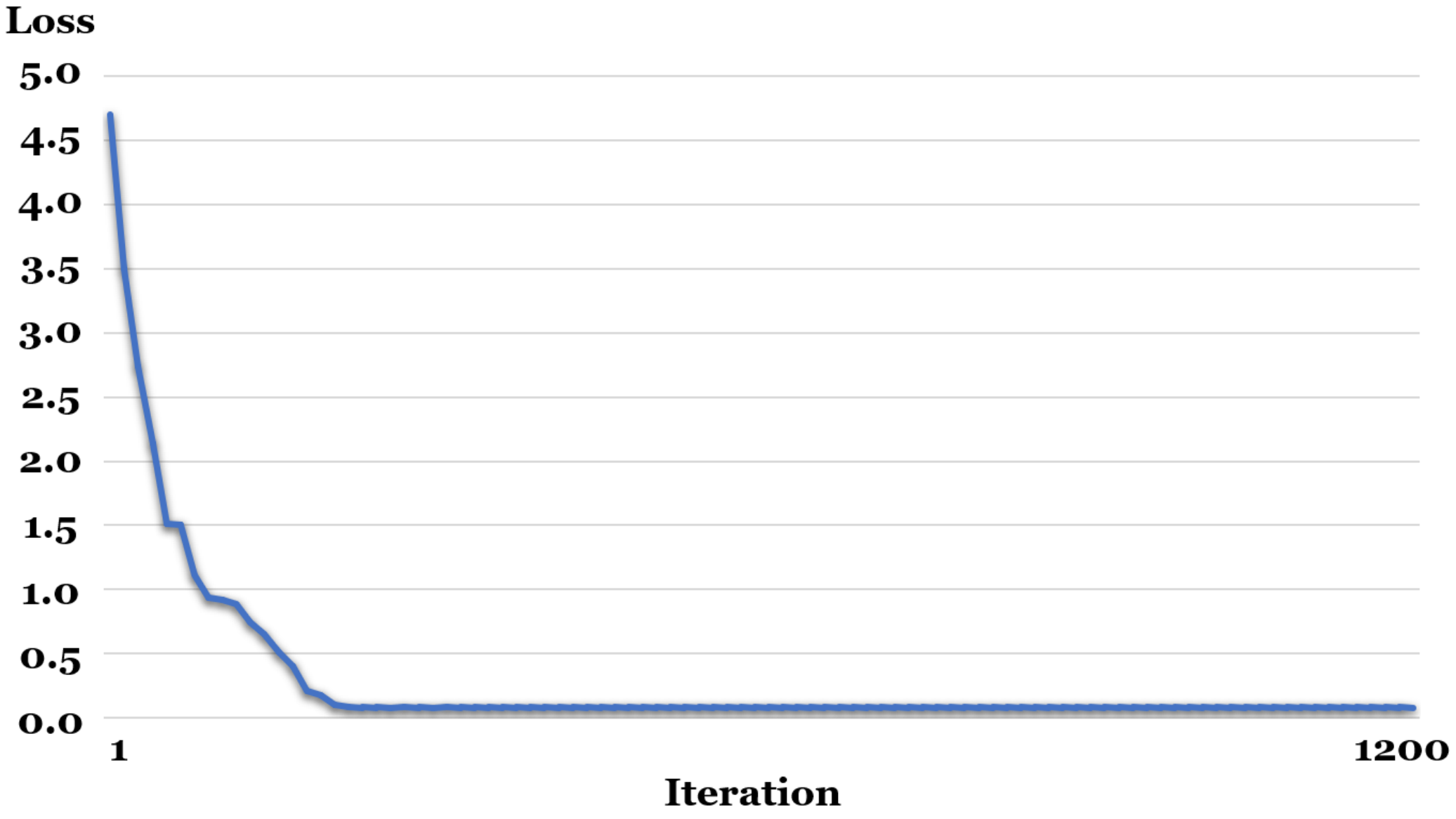

**Figure 10.** Graph showing losses by iteration

As a result of the model's learning, the total loss value was 0.03624. As an indicator for the evaluation of the learned model, accuracy was used. The method to derive is as follows:

$$Accuracy = \frac{The\ number\ of\ detected\ objects}{The\ number\ of\ total\ objects}$$

The building elevation detection accuracy and window detection accuracy were 0.93, 0.91, respectively.

*4.3 The result of extracting building elevation information based on image processing techniques*

This section describes the results of extracting the building elevation information using image processing techniques from the recognized building elevation images in the utility of a deep learning model. The current study divided building elevation information into 1) the simple classification method for windows, 2) the detailed classification method for windows, and 3) simple classification for walls. As described in Section 4.2, this section outlines the results of the detailed classification method for windows and the simple classification method for walls.

First, the detailed classification of windows focsued on the types of elevation windows (front curtain walls, repeated single windows, other) and the classificaiton of elevation window ratio (~25%, ~50%, ~75%, ~100%). To verify the suggested method, 43 front curtain wall elevation images, 31 repeated single window elevation images, and 38 other images (not front curtain wall or repeated single windows) were tested as test data. As a result, the accuracy in recognizing front curtain wall, repeated single windows, and other building types was 0.95 (41 images), 0.90 (28 images), and 1.00 (38 images), respectively.

Next, the simple classification for walls focused on the RGB channel value of the walls. After we removed the window areas from the elevation areas prior to extracting the RGB channel values, the average of the RGB channel values of the remaining area was calculated. "Matching" was performed based on the RGB channel value, which was calculated according to six previously defined RGB channel values. As a result, all 30 images for each color were well matched. This shows that the color of the building can be distinguished when RGB channel values are used. This methodology can be used in preprocessing for the detailed classification method for walls in future studies.

## 4. Conclusion

The current study used deep learning and image processing techniques to detect the elevation of buildings and extract information regarding their elevation. Subsequently, the techniques were 93% and 91% accurate in correctly detecting the elevation and windows, respectively, while showing excellent performance in experiments in which elevation information was extracted. However, in the process of detecting elevation and windows using deep learning techniques, the false detection rate during learning was higher than that of building elevation detection due to unclear window boundaries. In fact, noise data are highly likely to be mixed due to experimenter errors when generating learning data using VIA software. If this problem is resolved to conduct learning via the creation of a more detailed window area, the results will be more satisfactory.

In addition, the reliability of the accuracy may be reduced due to the small amount of data when the experimenter extracts building elevation information based on processing techniques. However, regarding information related to building elevation verifying performance with only a small amount of data is sufficient as it is recognized based on rules rather than the use of a learning model or statistical methodology. Moreover, this can be used in preprocessing for determining specific information such as the elevation material of a building in the future.

In summary, this study aimed to detect the elevation of a building using deep learning and image processing techniques to recognize and classify elevation information such as wall colors and the window ratio of the building. The ways in which the study contributes to the field are threefold. First, a model was built to learn the elevation and window images of the building using a 3D building elevation map, which has never previously been attempted. The current study acquired data by automating the 3D building elevation map extraction process, while previous studies were conducted with actual images in the form of street views. Unlike street views that were photographed from limited angles, this model has the advantage of being able to learn more sophisticated building elevation images from several angles. Second, the current study acquired results with better detection accuracies using the latest object detection model. Previous studies used basic image processing techniques or deep learning models that are not optimized for object detection, whereas this study used Fast R-CNN models that are optimized for object detection and segmentation within images. Third, a methodology for recognizing and extracting information regarding the elevation of buildings using image processing techniques was proposed in the current study. Future studies could adopt the proposed methodologies to acquire fundamental building elevation information automatically. Moreover, they can be used in preprocessing to recognize more detailed information regarding building elevation.

In using this methodology, it is possible to build information regarding exteriors of buildings while saving time, cost, and effort on a national level. The utility of information will be much more valuable via combining it with public data such as building utility ledgers. In particular, when the information related to building exteriors is constructed as a dataset on the level of BIM, the beauty of urban architecture, regional characteristics, city images, or landscape can be used as data, and a 'digital twin' can be implemented to build real-time information with the development of smart architecture and urban technology.

**Author Contributions:** Conceptualization and methodology, D.S. and N.B.; software and formal analysis and data curation, B.N.; writing—original draft preparation, D.S.; writing—review and editing, N.B.; visualization and supervision, D.S.; funding acquisition, D.S. and B.N. All authors have read and agreed to the published version of the manuscript.

**Funding:** This work was supported by the research grant of the Chungbuk National University in 2020.

**Acknowledgments:** In this section, you can acknowledge any support given which is not covered by the author contribution or funding sections. This may include administrative and technical support, or donations in kind (e.g., materials used for experiments).

**Conflicts of Interest:** The authors declare no conflict of interest.

## References

1. Korean Law Information Center. 「ACT ON PROMOTION OF THE PROVISION AND USE OF PUBLIC DATA」. Available online: https://www.law.go.kr/ (accessed0u on 2021.12.15).
2. Public Data Web Portal. List of Data. Available online: https://www.data.go.kr/tcs/dss/selectDataSetList.do (accessed on 2021.12.15).
3. Vworld Map. Available online: https://vworld.kr/v4po_main.do (accessed on 2021.12.15).
4. Seoul City 3D Map. S-MAP. Available online: https://smap.seoul.go.kr/ (accessed on 2021.12.15).
5. Gulshan, V.; Peng, L.; Coram, M.; Stumpe, M.C.; Wu, D.; Narayanaswamy, A.; Venugopalan, S.; Widner, K.; Madams, T.; Cuadros, J.; Kim, R.; Raman, R.; Nelson, P.C.; Mega, J.L.; Webster, D.R. Development and Validation of a Deep Learning Algorithm for Detection of Diabetic Retinopathy in Retinal Fundus Photographs. *JAMA* **2016**, *316(22)*, 2402-2410.
6. Lunit insight. https://insight.lunit.io/ (accessed on 2021.12.15).
7. Huang, W.; Song, G.; Hong, H.; Xie, K. Deep Architecture for Traffic Flow Prediction: Deep Belief Networks With Multitask Learning. *IEEE Transactions on Intelligent Transportation Systems* **2014**, *15(5)*, 2191-2201.
8. Huang, H.; Tang, Q.; Liu, Z. Adaptive Correction Forecasting Approach for Urban Traffic Flow Based on Fuzzy -Mean Clustering and Advanced Neural Network. *Journal of Applied Mathematics* **2013**, 2013, 7.
9. Noh, B.; No, W.; Lee, J.; Lee, D. Vision-Based Potential Pedestrian Risk Analysis on Unsignalized Crosswalk Using Data Mining Techniques. *Appl. Sci.* **2020**, 10, 1057. https://doi.org/10.3390/app10031057
10. Koo, W.; Yokota, T.; Takizawa, A.; Katoh, N. Image Recognition Method on Architectural Components from Architectural Photographs Glass Openings Recognition Based on Bayes Classification. *Architectural Institute of Japan* **2006**, 4, 123-128. (in Japanese)
11. Seo, D. study on the method for visual perception of architectural form through digital image processing. Doctorate Thesis, Yonsei University, Seoul, 2013. (in Korean)
12. Talebi, M.; Vafaei, A.; Monadjemi, A. Vision-based entrance detection in outdoor scenes. *Multimed Tools Appl* **2018,** 77, 26219–26238. https://doi.org/10.1007/s11042-018-5846-3
13. Armagan, A.; Hirzer, M.; Roth, P. M.; Lepetit, V. Accurate Camera Registration in Urban Environments Using High-Level Feature Matching. British Machine Vision Conference (BMVC), London, UK, 2017.9.2.
14. Seong, H.; Choi, H.; Son, H.; Kim, C. Image-based 3D Building Reconstruction Using A-KAZE Feature Extraction Algorithm. 34th International Symposium on Automation and Robotics in Construction, 2018.3.
15. Yuan, L.; Guo, J.; Wang, Q. Automatic classification of common building materials from 3D terrestrial laser scan data. *Automation in Construction* **2020**, 110, 103017.
16. YOLO. You Only Look Once: Unified, Real-Time Object Detection. Available online: https://www.arxiv-vanity.com/papers/1506.02640/ (accessed on 2021.12.15).
17. AUTOIT. Available online: https://www.autoitscript.com/site/ (accessed on 2021.12.15).
18. VGG Image Annotator (VIA). Available online: https://www.robots.ox.ac.uk/~vgg/software/via/via.html / (accessed on 2021.12.15).
19. Szegedy, C.; Toshev, A.; Erhan, D. Deep Neural Networks for Object Detection. Advances in Neural Information Processing Systems. 1-9. 2013.12.5.
20. Purkait, P.; Cheng, Z,; Christopher, Z. SPP-Net: Deep Absolute Pose Regression with Synthetic Views. British Machine Vision Conference(BMVC 2018). 2017.12.9.
21. Wang, L.; Guo, S.; Huang, W.; Qiao, Y. Places205-VGGNet Models for Scene Recognition. ArXiv abs/1508.01667. 2015.8.7.

22. Girshick, R.; Donahue, J.; Darrell, T.; Malik. J. Rich feature hierarchies for accurate object detection and semantic segmentation. Proceedings of the IEEE conference on computer vision and pattern recognition. 2014.10.22.
23. Girshick, R. Fast r-cnn. Proceedings of the IEEE international conference on computer vision. 2015.9.27.
24. Ren, S.; He, K.; Girshick, R., Sun, J. Faster R-CNN: towards real-time object detection with region proposal networks. IEEE transactions on pattern analysis and machine intelligence. 2016.1.6.
25. Tian, Y.; Yang, G.; Wang, Z.; Wang, L.; Liang, Z. Apple detection during different growth stages in orchards using the improved YOLO-V3 model. *Computers and electronics in agriculture* **2019**, 157, 417-426.
26. Shafiee, M.; Chywl, B.; Li, F.; Wong, A. Fast YOLO: A fast you only look once system for real-time embedded object detection in video. arXiv preprint arXiv:1709.05943. 2017.9.18.
27. Redmon, J.; Farhadi, A. Yolov3: An incremental improvement. arXiv preprint arXiv:1804.02767. 2018.4.8.
28. Bochkovskiy, A.; Wang, C.; Liao, H. Yolov4: Optimal speed and accuracy of object detection. arXiv preprint arXiv:2004.10934. 2020.4.23.
29. He, K.; Gkioxari, G.; Dollar, P.; Girshick, R. Mask r-cnn. Proceedings of the IEEE international conference on computer vision. 2018.1.24.
30. Pham, V.; Pham, C.; Dang, T. Road Damage Detection and Classification with Detectron2 and Faster R-CNN, 2020 IEEE International Conference on Big Data (Big Data), 2020, pp. 5592-5601, doi: 10.1109/BigData50022.2020.9378027
31. Detectron2. Available online: https://github.com/facebookresearch/detectron2 (accessed on 2021.12.15).
32. Chernov, V.; Alander, J.; Bochko, V.; Integer-based accurate conversion between RGB and HSV color spaces, *Computers & Electrical Engineering* **2015**, 46, 328-337.